# Predicting Subjective Features of Questions of QA Websites using BERT

Issa Annamoradnejad  
Department of Computer Engineering  
Sharif University of Technology  
Tehran, Iran  
moradnejad@ce.sharif.edu

Mohammadamin Fazli  
Department of Computer Engineering  
Sharif University of Technology  
Tehran, Iran  
fazli@sharif.edu

Jafar Habibi  
Department of Computer Engineering  
Sharif University of Technology  
Tehran, Iran  
jhabibi@sharif.edu

*Abstract*— Community Question-Answering websites, such as StackOverflow and Quora, expect users to follow specific guidelines in order to maintain content quality. These systems mainly rely on community reports for assessing contents, which has serious problems such as the slow handling of violations, the loss of normal and experienced users' time, the low quality of some reports, and discouraging feedback to new users. Therefore, with the overall goal of providing solutions for automating moderation actions in Q&A websites, we aim to provide a model to predict 20 quality or subjective aspects of questions in QA websites. To this end, we used data gathered by the CrowdSource team at Google Research in 2019 and a fine-tuned pre-trained BERT model on our problem. Based on the evaluation by Mean-Squared-Error (MSE), the model achieved a value of 0.046 after 2 epochs of training, which did not improve substantially in the next ones. Results confirm that by simple fine-tuning, we can achieve accurate models in little time and on less amount of data.[1]

## I. Introduction

In recent years, online Q&A websites have attracted many users and are considered as reliable sources by experts from various fields. These websites provide an interface for users to exchange and share knowledge. The user asking a question lacks knowledge of a specific topic and experts on the same topic provide the desired knowledge. In this way, questions and answers are the source of information, replacing other sources like documents or databases [1].

These systems, in addition to general user rules, have specific rules to maintain their content quality. Question quality is essential both for the personal use of users and for the question and answering platforms as a whole, because high-quality question-answer pairs attract more users and improve platform traffic [2]. Therefore, detecting, removing or editing low-quality questions is a necessary step for the success of websites.

Due to the vast expanse of some of these systems in terms of the number of users, manual check and verification of new content by the administrators and official moderators are not feasible, and these systems require scalable solutions. In major Q&A networks, such as Stack Exchange websites[2], the current strategy is to use crowdsourcing and reliance on user reports. This strategy has serious problems, including the slow handling of violations, the loss of normal and experienced users' time, the low quality of user reports, and discouraging feedback to new users.

With the overall goal of providing solutions for automating moderation actions in Q&A websites, in this research, we plan to provide a model that could predict 20 quality or subjective aspects of questions. Since these aspects include questions about opinions, recommendations, or personal experiences, they are harder to answer by computer than questions with single, verifiable answers [3].

Given the need to maintain quality standards for the contents of online Q&A communities and the significant problems of crowdsourcing, providing solutions and models for automatically detecting user violations can bring upon faster detection of user violations (such as detection of spam, advertisement, grammar faults, etc) and therefore, saving users time and improve the quality of the contents.

Unfortunately, it is hard to build subjective question-answering algorithms, because of a lack of data and predictive models. In this article, we use data gathered by the CrowdSource team at Google Research, a group dedicated to advancing NLP and other types of ML science via crowdsourcing [3-5]. We used this new data set to build predictive algorithms for different subjective aspects of question quality.

Transfer Learning allowed researchers to smash multiple benchmarks with minimal task-specific fine-tuning and provided the rest of the NLP community with pre-trained models that could easily with less data and less compute time be fine-tuned to produce state of the art results [6]. Google's BERT is one of these models that theoretically allows us to smash multiple benchmarks with minimal task-specific fine-tuning [7].

---

[1] Our code is available at: https://github.com/Moradnejad/Predicting-Subjective-Features-on-QA-Websites

[2] A group of 170+ QA websites in diverse fields covering specific topics, such as Stackoverflow.com, Askubuntu.com and Superuser.com.

Therefore, this paper aims at predicting different subjective aspects of questions in question-answering websites using BERT. Since the task is to predict values of 20 target qualities of questions, which they are all related to the question title and body, therefore, those that relate to the answer feature are excluded from this research.

The structure of this article is as follows: Section 2 reviews past works on quality prediction of questions in QA websites and latest NLP models. Section 3 explains the data and section 4 elaborates on the methodology. In Section 5, evaluation of the experiment is presented, and Section 6 is the concluding remarks.

## II. LITERATURE REVIEW

A few works focused on quantifying the quality of a question by analyzing their features. Research has shown that too short questions have a low probability of obtaining an answer [8]. Another study analyzed unanswered questions in Stackoverflow and found that answered questions have higher scores compared to unanswered questions [9].

In [10] authors find that the number of answers is the most significant feature to predict the long-term value of a question together with its answers set and it is direct feedback on the usefulness/quality of the question. However, this proposed feature cannot be used on new questions since they are yet to be answered.

Transfer learning, particularly models like ULMFiT, Allen AI's ELMO, and Google's BERT, focuses on storing knowledge gained from training on one problem and applying it to a different but related problem (usually after simple fine-tuning on small amount of data).

The first transfer learning method in Natural Language Processing (NLP) was Universal Language Model Fine-tuning for Text Classification (ULMFiT) method [11] that fine tunes the language model on a new data set after training a language model on a preliminary data set, such as Wikitext. Finally, the resulting fine-tuned language model can be used in a prediction task for a new data set. The model, besides significantly outperforming many state-of-the-art tasks, was done by training on only 100 labeled examples, that matched performances equivalent to old models trained on 100× more data.

ELMo is another related study that includes task-specific architectures and uses the pre-trained representations as additional features [12]. It is a deep contextualized word representation that represented complex characteristics of word use (e.g., syntax and semantics).

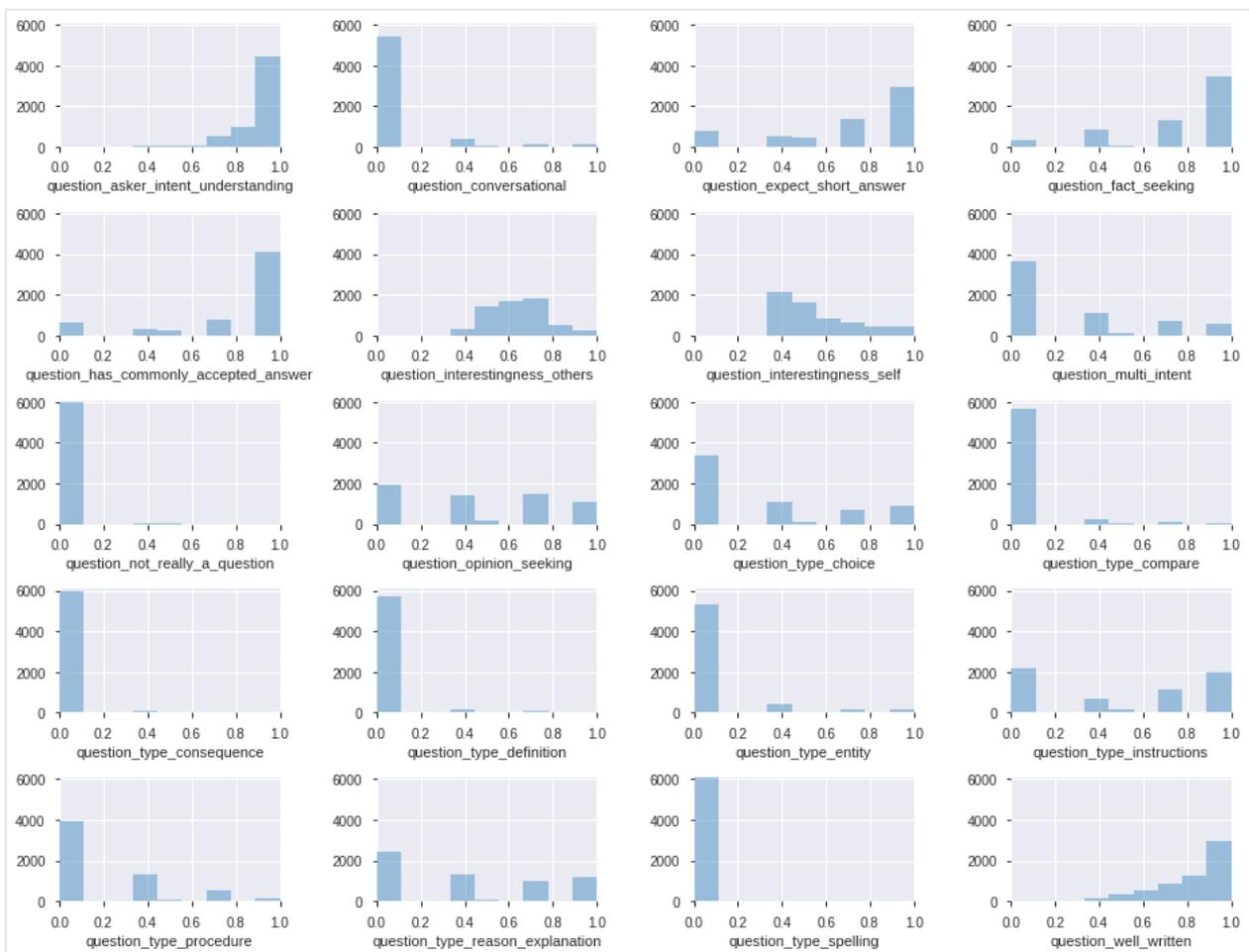

Figure 1 Distribution of 21 target values. Values between 0 and 1; total sample size=6079.

The BERT language model consists of several transformer encoders stacked together and is designed from unlabeled text to pretrain deep bidirectional representations by jointly conditioning on left and right context in all layers [7]. They presented two general types, named the BERT-base and the BERT-large, which used the BooksCorpus with 800M words [13] and English Wikipedia with 2,500M words [14], respectively. The characteristics of these models are presented below (L is the number of stacked encoders, H the hidden layer size, and A the number of self-attention heads):

1. BERT-base: L:12, H: 768, A:12, Parameters: 110M
2. BERT-large: L:24, H:1024, A:16, Parameters: 340M

Table 1 Name and description of target columns

|   | Target Column | Column description |
|---|---|---|
| 1 | asker intent understanding | Is the question's intent understood well? |
| 2 | conversational | Is the question conversational? |
| 3 | expect short answer | Does the question expect short answer? |
| 4 | fact seeking | Is the question looking for factual information? |
| 5 | has commonly accepted answer | Does the question has commonly accepted answer? |
| 6 | interestingness others | Does the question look interesting to others? |
| 7 | interestingness self | Does the question look interesting to the asker? |
| 8 | multi intent | Does the question has multiple intents (multiple questions inside one)? |
| 9 | not really a question | Should the question be reported as not a question? |
| 10 | opinion seeking | Does the question asks for opinion-based answers? |
| 11 | type choice | Is the question a multi-choice question? |
| 12 | type compare | Is the question looking for comparison between alternatives? |
| 13 | type consequence | Is the question looking for consequence of a possible action? |
| 14 | type definition | Is the question looking to define something? |
| 15 | type entity | Is the question related to an entity? |
| 16 | type instructions | Is the question looking for instructions? |
| 17 | type procedure | Is the question about a procedure/looking for procedure? |
| 18 | type reason explanation | Is the question looking expects explanation of reason? |
| 19 | type spelling | Is the question about spelling? |
| 20 | well written | Is the question well-written? |

III. DATA

In this section, we will briefly explain our used data set, alongside some general statistics on the data set.

*A. Data*

The CrowdSource team at Google Research, a group dedicated to advancing NLP and other types of ML science via crowdsourcing, has collected data on a number of quality scoring aspects of questions and answers from QA websites in 2019. The questions consists of 5 categories: "Technology", "Stackoverflow", "Culture", "Science", "Life arts" and was collected from nearly 70 different stack exchange websites. Reportedly, raters received minimal guidance and training, and relied largely on their subjective interpretation of the prompts. As such, each prompt was crafted in the most intuitive fashion so that raters could simply use their common sense to complete the task [3].

Data is relatively small, only made of 6079 rows. It originally contains 40 columns, of which 10 are given as basic features (question title, body, answer …) and the rest are target quality labels. Of the 10 feature columns, we only used 'question_title' and 'question_body' for the purposes of this study and the rest is excluded. From the 30 target columns, 21 are related to the question and the remaining 9 columns are related to answer quality. Thus, we excluded columns related to answer from the data set. Finally, we removed 'question_body_critical' column, as its intent and meaning was not clear to us and no definition was given.

Table 1 gives the name and description of the selected target columns that we focus in this research. As we said earlier, target labels are aggregated from multiple raters and have continuous values in the range [0,1].

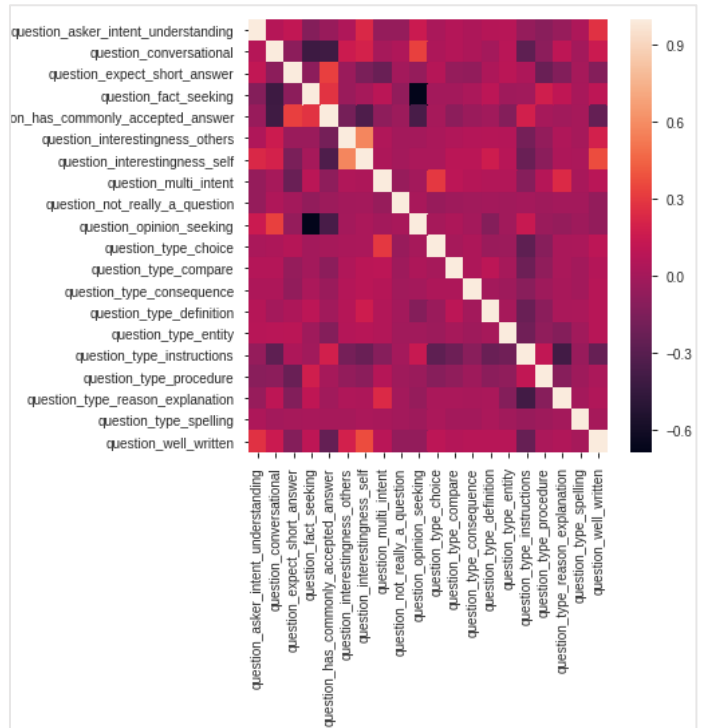

Figure 2 Correlation heat-map for the target columns

## B. General Statistics and EDA

In this section, we introduce a few general statistics on the data set.

Figure 1 depicts distribution of the selected target values. Since the values are real values, we used 'numpy.linspace' to create 10 sub-ranges (0.1 each) for the purposes of this depiction. We can see that most of the columns, such as 'question_asker_intent_understanding' and 'question_conversational', are unevenly distributed and aggregated in one side value (0 or 1).

Figure 2 depicts correlation heat-map for the values of the target columns. We can barely view any strong correlation (lighter color) between the values of these columns. However, we can see a few anti-correlations (dark colors), such as conversation vs. fact, conversational vs. has commonly accepted answer, and fact-seeking vs. opinion-seeking.

Even though, because of our pre-trained model we do not require to perform feature extraction/engineering, we extracted a few simple features and depicted their correlation with target values. This was done to better understand the underlying latent relationships in data. To this end, we extracted character count and word count on question title and body, punctuations count in question body, and number of duplicate words, duplication rate, and number of sentences in question body. Correlation of these extracted features with target columns are depicted in Figure 3. We can see that the extracted features do not have string correlations with the target values (coefficient between -0.17 to +0.25).

Finally, we performed sentiment analysis on the question body to extract polarity and subjectivity of texts. The mean for polarity is positive, while the questions are prone to be objective. Scatter plot for this analysis is depicted in Figure 4.

## IV. PROPOSED METHOD

We used transfer learning from pre-trained transformers. Ref [7] introduced a new language representation model called BERT, which was described partially in Section II. BERT stands for Bidirectional Encoder Representations from Transformers and is a multi-layer bidirectional Transformer encoder. The architecture is designed to pre-train deep bidirectional representations from unlabeled text by jointly conditioning on both left and right context in all layers.

In using pre-trained BERT model, the steps are: (1) plug in the task specific inputs and outputs into BERT and (2) fine-tune all the parameters end-to-end. Therefore, by fine-tuning pre-trained BERT model we can get away with traditional training. Our instance loops over the folds in GroupKFold and trains each fold for 5 epochs with a batch_size of 6. A binary crossentropy is used as the objective loss function. We did not alter any of the BERT default parameters; however, we did perform the fine-tuning on our data for learning rate values between 1e-5 to 9e-5.

We used 'tensorflow' and 'huggingface transformers' libraries in python for our base model, and *sklearn, pandas, numpy, matplotlib* and *math* libraries for data loading, additional preprocessing, and evaluation metrics.

HuggingFace created the well-known transformers library, formerly known as 'pytorch-transformers' or 'pytorch-pretrained-BERT', it brings together over 40 state-of-the-art pre-trained NLP models (BERT, GPT-2, RoBERTa, CTRL…) [15]. The implementation gives interesting additional utilities like tokenizer, optimizer or scheduler that we utilized in the implementation of this paper. While the transformers library can be self-sufficient, but incorporating it within the 'fastai' library provides simpler implementation compatible with powerful fastai tools like Discriminate Learning Rate, Gradual Unfreezing or Slanted Triangular Learning Rates.

## A. Preprocessing

We performed the default model specific tokenizers provided by 'huggingface'. For target variables, first we applied rank transform method (values are replaced with their corresponding ranks) and then, we continued with min-max scaling. This made the target values evenly distributed between 0 and 1.

We defined the maximum sequence length that will be used for the input to BERT (maximum is usually 512 tokens). Due to the 2 x 512 input, it will require significantly more memory when fine-tuning BERT.

## V. EVALUATION

We used 80% of data for training and 20% (1216 rows) for validation. Mean squared error was utilized on the validation set to gain the accuracy of the model. Based on our results (Table 2) on 4864 rows, we achieved value of 0.05 for MSE in all of learning rate values after one epoch of training. After three epochs, model achieved values between 0.046 and 0.048, which did not improve in the next epochs. Our best results came from the second iteration with LR=3e-5 and LR=5e-5, where the model achieved MSE with a value of 0.046. In conclusion, we can say that epochs of training did not substantially improve model's accuracy and one epoch of training is enough to achieve an accurate model.

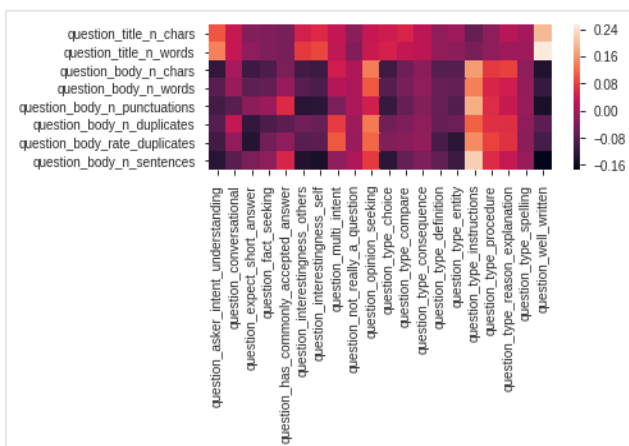

Figure 3 Correlation heat-map between extracted sample features and target values

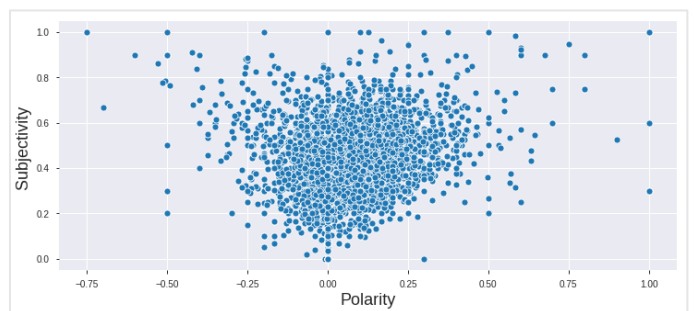

Figure 4 Results of sentiment analysis on question body

Table 2 Accuracy of our model with different LRs in 5 epochs

| Epoch | LR=1e-5 | LR=3e-5 | LR=5e-5 | LR=7e-5 | LR=9e-5 |
|---|---|---|---|---|---|
| 1 | 0.055 | 0.051 | 0.051 | 0.051 | 0.052 |
| 2 | 0.052 | **0.046** | **0.046** | 0.05 | 0.05 |
| 3 | **0.048** | 0.048 | 0.049 | **0.049** | **0.049** |
| 4 | 0.049 | 0.051 | 0.054 | 0.052 | 0.053 |
| 5 | 0.049 | 0.053 | 0.052 | 0.051 | 0.053 |

In case of timing and performance, it took 8.76 minutes in average for each epoch on a computer with 16GB RAM and Intel(R) Xeon(R) CPU 2.00GHz.

## VI. CONCLUSION

In this study, we experiment the BERT model to solve a part of the problem of moderation actions in QA websites. To be specific, we used BERT to predict 20 quality or subjective aspects of questions in QA websites. Predicting subjective aspects is a hard problem with computers and we showed that it could benefit from transfer learning from pre-trained transformers.

Our model achieved MSE with a value of 0.046 in predicting target values. Results confirm that by simple fine-tuning pre-trained BERT model, we can achieve high accuracy, in little time, and on less amount of data.